\documentclass[10pt,twocolumn,letterpaper]{article}
\pdfoutput=1

\usepackage{cvpr}
\usepackage{times}
\usepackage{graphicx}
\usepackage{amsmath}
\usepackage{amssymb}

\usepackage{array}
\usepackage{kbordermatrix}
\usepackage{booktabs}
\usepackage{enumitem}
\usepackage{microtype}
\usepackage{textcomp}

\usepackage[pagebackref=true,breaklinks=true,letterpaper=true,colorlinks,bookmarks=false]{hyperref}
\usepackage[capitalize,noabbrev]{cleveref} 

\cvprfinalcopy 

\def\cvprPaperID{8410} 

\ifcvprfinal\fi

\DeclareMathOperator{\Span}{Span}
\DeclareMathOperator{\SO}{SO}

\newcommand{\rot}[1]{\mathtt{#1}}
\newtheorem{thm}{Theorem}

\newtheorem{defn}{Definition}

\begin{document}

\title{On the Distribution of Minima in Intrinsic-Metric Rotation Averaging}

\author{Kyle Wilson\\
Washington College\\ Department of Math and Computer Science \\
Chestertown, MD\\
{\tt\small kwilson24@washcoll.edu}
\and
David Bindel\\
Cornell University \\ Department of Computer Science\\
Ithaca, NY\\
{\tt\small bindel@cs.cornell.edu}
}

\maketitle
\thispagestyle{empty}

\begin{abstract}
   
Rotation Averaging is a non-convex optimization problem that determines
orientations of a collection of cameras from their
images of a 3D scene.
The problem has been studied using a variety of distances and
robustifiers.
The intrinsic (or geodesic) distance on $\SO(3)$
is geometrically meaningful; but while some extrinsic distance-based
solvers admit (conditional)
guarantees of correctness, no comparable results
have been found under the intrinsic metric.

In this paper, we study the spatial distribution of local minima. First, we do
a novel empirical study to demonstrate sharp transitions in
qualitative behavior: as
problems become noisier, they transition from a single (easy-to-find) dominant
minimum to a cost surface filled with minima. In the second part of this paper we
derive a theoretical bound for when this transition occurs. This is an
extension of the results of~\cite{wilson-eccv-16}, which used
local convexity as a proxy to study the difficulty of problem.
By recognizing the underlying quotient manifold geometry of
the problem we achieve an $n$-fold improvement over prior
work. Incidentally, our analysis also extends the prior $l_2$ work to
general $l_p$ costs. Our results suggest using algebraic connectivity
as an indicator of problem difficulty.

\end{abstract}

\section{Introduction}
\label{sec:intro}

The rotation averaging problem arises in computer vision as part of
estimating the 3-dimensional poses of a set of
cameras.
We are
given many images of some scene,
produced by cameras at unknown locations and with unknown orientations.
When two images
overlap,
it is often possible
to estimate
the
\emph{relative orientation} of the associated cameras.
By considering many overlapping camera pairs we obtain many such
pairwise measurements.

Essentially,
rotation averaging is
an optimization problem on a
graph in which vertices represent cameras and edges represent
measurements of relative orientation.
The goal
is to pick an \emph{absolute orientation} for each vertex
in a way that best agrees with the
relative orientation measurements.
Both the relative
and absolute orientations must be in
$\SO(3)$, the group of orientation-preserving rotations of 3D space.

We may choose the cost function used in rotation averaging
for convenience of
computation, for the existence of theoretical guarantees, or for
modeling concerns. From a geometric perspective,
costs
based
on intrinsic metrics---geodesic distances on $\SO(3)$---are
a natural choice.
However, no such method is
presently known to have a guarantee of global convergence. In fact,
the $l_2$ geodesic cost is known to yield a non-convex problem.
Formulations based on
extrinsic costs---non-manifold distances within representations---have
been more tractable and
sometimes give optimality guarantees.
When solvers fail the junk solutions doom downstream problems, such as translations averaging and bundle adjustment.

This paper considers cost functions based on the intrinsic, geodesic
distance.
Two recent surveys~\cite{carlone-icra-15,tron-cvprw-16} propose
a familiar pattern:
use an
extrinsic solver to
generate a candidate solution,
then refine with
with a solver
chosen for its modeling properties---even if the latter solver only
guarantees local optimality. This is a common pattern in geometric
computer vision, with a well-known exemplar being the
``Gold-Standard'' algorithm
for estimating homographies~\cite{HZ}.
The extrinsic-intrinsic solver pattern leads to some key questions:
``Is this initial guess good enough
for us to discover the best solution?'' and ``Why do some
problems require much better initial guesses than others?'' In this
paper, we seek partial answers to these questions.
(Note that we do not propose a new solver.)
We make two main contributions:
\begin{enumerate}
    \item We empirically investigate the cost surfaces of intrinsic rotation
    averaging problems. We
    demonstrate how properties of the problem instance can lead to
    challenging distributions of local minima.
    \item We derive bounds on this behavior by improving the local convexity
    analysis in~\cite{wilson-eccv-16}. We show how rotation averaging's gauge symmetry leads to a
    quotient-manifold description of the optimization problem, which
    leads to an $n$-fold improvement in bounds.

\end{enumerate}

\section{Related Work}
\label{sec:related}

\textbf{Relaxed Approaches.}
The seminal paper introducing the
rotation averaging problem~\cite{govindu-cvpr-01} also
proposed a solver based on the unit quaternion representation of 3D
rotations.
This is a prototypical \emph{relaxed} method: by ignoring the length-1
constraint and minimizing Euclidean distances between quaternions, the
problem becomes linear.
The solution may not satisfy the
unit constraint (hence not actually represent rotations),
but it can be normalized easily.
Many later methods follow the same pattern of solving a relaxed
problem, then ``rounding'' the solution to satisfy the constraints.
Most works in this line use a rotation matrix representation and relax
the orthnormality and determinant constraints. Martinec and
Pajdla~\cite{martinec-cvpr-07} relaxed to a least-squares
problem. Arie-Nachimson \etal~\cite{arienachimson-3dimpvt-12} give a
spectral solution to a similar
cost. Arrigoni \etal~\cite{arrigoni-3dv-14} solve their relaxed
problem in a low rank + sparse framework for greater robustness to
outliers.  Wang and Singer~\cite{wang-ii-13} seek robustness by
solving an $l_1$ relaxed problem.
Unfortunately,
even if the unconstrained problem can be solved exactly, the final
rounded solution is generally not an optimal solution to any
particular problem, and the ``relaxation gaps'' due to rounding are not
well understood.

\textbf{Manifold Approaches.}
Manifold methods
use the tools of Riemannian geometry to optimize directly on the
rotations manifold.
In practice, this involves iteratively solving a
series of Euclidean tangent space problems.
Govindu~\cite{govindu-cvpr-04} pioneered this approach, and
Hartley \etal~\cite{hartley-ijcv-13} and Chatterjee and
Govindu~\cite{chatterjee-pami-17} have proposed robustified
methods. Tron \etal~\cite{tron-icdsc-08} give a distributed consensus
solver for a specific ``reshaped'' cost.
Unfortunately, while such solvers converge to a local minimum,
that local minimum may not be globally optimal.
Two survey
papers~\cite{carlone-icra-15,tron-cvprw-16} propose using an
initialize-and-refine pattern, where a globally solveable relaxed
method generates an initial guess which a manifold-based method then
refines.

\textbf{Verification and Guarantees.}
More recently, there has been particular interest in validation and
performance guarantees for solvers. Fredriksson and
Olsson~\cite{fredriksson-accv-12} return to quaternions, and are able
to verify---via Lagrangian Duality---that their solution is optimal
(if noise levels are low enough). Briales and
Gonzales-Jimenez~\cite{briales-iros-16} give a verifier for camera
poses, rather than
rotations.
Boumal \etal~\cite{boumal-dc-13,boumal-ii-14} take a
statistical approach
and study the efficiency of maximum
likelihood estimators via Cram\'er-Rao bounds.  They develop
visualizations to show how the problem's graph topology affects the
uncertainty in the output.
In contrast, this paper visualizes how topology affects the
macro-scale shape of the cost surface, independent of a choice of solution.

Most recently,
Ericsson \etal~\cite{eriksson-cvpr-18} propose a new solver based on
Lagrangian duality. Unusually, this method uses an extrinsic cost but
makes no relaxations (and hence suffers no rounding gap). When strong
duality holds, this method gives a global solution. For complete
graphs, they show that it is sufficient that residuals are less than
$42.9^\circ$.
We find it interesting that two very different approaches (theirs: chordal distance and Lagrangian duality and ours: geodesic distance and local convexity) lead to statements with similar form.
Briales and Gonzales-Jimenez~\cite{briales-ral-17} and Rosen \etal's~\cite{rosen-ijrr-19}
fast relaxation methods also give verifiers.

This present paper is most closely related to Wilson \etal~\cite{wilson-eccv-16} which studied $l_2$ intrinsic costs via a local convexity analysis. It concluded that some rotation averaging problem instances exhibit large areas of convex behavior, but their extent is related to problem size, noisiness, and connectivity. However, the local convexity is not apparent until an underlying gauge ambiguity is removed. This paper returns to the analysis in~\cite{wilson-eccv-16} with a more geometrically natural description of gauge which yields greatly improved bounds. The present analysis is also more general, as it considers an $l_p$ cost rather than the $l_2$ cost of~\cite{wilson-eccv-16}. Also, unlike~\cite{eriksson-cvpr-18} and \cite{wilson-eccv-16}, we empirically investigate what happens outside the optimality bounds.

\textbf{Outline of Remaining Sections.}
The remainder of this paper proceeds as follows: \Cref{sec:prelim} gives notation and background on representations of rotations. It also states the rotation averaging problem in the form we will use. \Cref{sec:empirical} is an empirical investigation into the shape of the cost surfaces of our problems. We use a spectral embedding to visualize the locations and strengths of local minima. These experiments clearly demonstrate transitions from easy problems dominated by a single minimum to difficult problems with a preponderance of bad local minima. In \cref{sec:theory} we bound this behavior by improving the local convexity analysis from~\cite{wilson-eccv-16}, and finally we suggest applications and conclude in \cref{sec:conclusion}.

\section{Preliminaries}
\label{sec:prelim}

In this section we introduce notation and formally define the rotation averaging problem. We will also note some properties which are readily apparent.

\subsection{3D Rotations and Their Representations}
The special orthogonal group in 3 dimensions---$\SO(3)$---is the group
of rotations
of $\mathbb{R}^3$. $\SO(3)$ is a smooth 3-dimensional manifold, so it
is in fact a Lie group---a notion of continuous symmetry.
While this suffices as a geometric definition, for computational
purposes we require coordinatized representations.
Euler's rotation theorem states that every rotation can be decomposed
(not uniquely) into an angle $\theta$ and an axis $\hat{\mathbf{v}}$.
Alternately, every element of $SO(3)$ is uniquely represented by
a $3 \times 3$ orthogonal matrix with
determinant 1.
As orthogonal matrices, rotation matrices have
the property that $\rot{R}^{-1} = \rot{R}^\top$.
The angle-axis representation $\theta \hat{\mathbf{v}}$ is closely related to the tangent space
to $\SO(3)$, which is 3-by-3 skew symmetric matrices. We write a tangent space element:
\[ [\theta \hat{\mathbf{v}}]_\times =
\left[ \begin{array}{ccc}
    0 & -\theta v_z & \theta v_y \\
    \theta v_z & 0 & -\theta v_x \\
    -\theta v_y & \theta v_x & 0 \\
\end{array} \right],
\;\;
\text{where} \;
\hat{\mathbf{v}} =
\left[ \begin{array}{c}
    v_x \\
    v_y \\
    v_z \\
\end{array} \right].
\]

Rotation matrices form a 3-dimensional submanifold embedded in $\mathbb{R}^{3\times3}$, giving rise to at least two common metrics:
\begin{align}
    d_{\mathrm{intrinsic}}(\rot{R}, \rot{S}) &= \angle(\rot{R}^\top \rot{S}) \\
    d_{\mathrm{extrinsic}}(\rot{R}, \rot{S}) &= \| \rot{R} - \rot{S} \|_F.
\end{align}
where $\angle(\cdot)$ is the angle of a given rotation (\emph{a la} Euler's rotation theorem) and $\|\cdot\|_F$ denotes the Frobenius matrix norm. The intrinsic metric can be shown to be a geodesic distance---that is, the length of the shortest path \emph{that follows the curvature of the rotation manifold}. The extrinsic metric cuts through  non-manifold space: it is a distance in $\mathbb{R}^{3\times3}$, and so is generally shorter than the $d_{\mathrm{intrinsic}}$. The two are related by $d_{\mathrm{extrinsic}} = 2 \sqrt{2} \sin \left( \frac{d_{\mathrm{intrinsic}}}{2} \right)$.
This paper is about rotation averaging under the intrinsic metric, so we will omit the subscripts and simply write $d_{\mathrm{intrinsic}}$ as $d$.

\subsection{Problem Definition}
A rotation averaging problem instance consists of a graph $G=(V,E)$ with
measurements
$\widetilde{\mathcal{R}} = \{ \widetilde{\rot{R}}_{ij} | (i,j) \in
E\}$ of relative orientation on each edge.
Writing $n=|V|$, our goal is to choose an absolute orientation
$\mathcal{R} \in \SO(3)^n$ for each vertex in a way that
best respects those measurements:
\begin{equation}
    \min_{\mathcal{R} \in \SO(3)^n}
    \phi_p(\mathcal{R}) = \sum_{(i,j) \in E} d(\, \widetilde{\rot{R}}_{ij}, \rot{R}_i \rot{R}_j^\top)^p.
    \label{eqn:problem}
\end{equation}
We will refer to $\phi_p$ as the $l_p$ cost function for this problem. Our results in \cref{sec:theory} will apply for the $p>1$ cases.

\subsection{Local Minima}
The $l_p$ problems in \cref{eqn:problem} are non-convex,
manifold-valued optimization problems.
Analogous problems in Euclidean space are convex;
the nonconvexity of our problem arises entirely
from the manifold geometry.
Iterative solvers of the type we consider start from an initial guess,
then iterate toward a stationary point (which we hope is a local minimum).
This local minimum need not be a global minimizer for the
problem, and \Cref{sec:empirical} shows that in some situations the
problem may have many bad local minima.
The guiding concern of the analysis in this paper is improving our
understanding of this failure mode.
Note that the multiplicity and spatial distribution of local minima are
properties of a problem instance itself, not of a solver.
In principle,
this difficulty
can be largely mitigated
with a
sufficiently high quality guess. Thus, it would be helpful to have
some theoretical guidance as to what quality of guess is required.

\subsection{Some Simple Observations}
{\bf Edge Directions.}
If $\widetilde{\rot{R}}_{ij}$ is a measurement of $\rot{R}_i \rot{R}_j^\top$---the relative rotation on edge $(i,j)$---then $\widetilde{\rot{R}}_{ij}^\top$ is a measurement of $\rot{R}_j \rot{R}_i^\top$. We will treat $G$ as an undirected graph, and supply either $\widetilde{\rot{R}}_{ij}$ or $\widetilde{\rot{R}}_{ij}^\top$ as appropriate in context.

{\bf Minimally Constrained Problems.}
We assume throughout that $G$ is connected.
When $G$ is a tree there always
exists a solution with zero cost. Any additional edges overconstrain
the problem,
but this redundancy improves accuracy in the face of measurement noise.

{\bf Gauge Ambiguity.}
Any rotation averaging cost function of the form above is right-invariant to a multiplication of the form
$\mathcal{R} \rightarrow \mathcal{R} \rot{S}$
which maps rotations
$\rot{R}_i \mapsto \rot{R}_i \rot{S}, \; i \in V$.
Given any point $\mathcal{R} \in \SO(3)^n$, the cost function is constant on the entire \emph{gauge-equivalent} set $\{ \mathcal{R} \rot{S}, \rot{S} \in \SO(3) \}$. The $3n$-dimensional problem domain can be considered to be partitioned into $3$-dimensional gauge orbits,
and solutions are
only unique up to the choice of orbit representative.
An extra {\em gauge-fixing} constraint can be used to pick the orbit
representative uniquely.
The simplest method of gauge-fixing is to pick one vertex arbitrarily
(say, vertex $k$), and
set $\rot{R}_k = \mathbf{I}_3$.
Later in the paper, we will consider an alternate approach that
factors out the symmetry. Interestingly, Briales and
Gonzalez-Jimenez~\cite{briales-iros-16} have
observed faster solver
performance when moving from simple vertex fixing to a scheme similar to ours.

\section{Some Empirical Observations}
\label{sec:empirical}
In this section, we demonstrate
a way to visualize the spatial distribution
of local minima in rotation averaging
problems. It shows us that structural factors (such as edge density
and connectivity) can greatly affect the distribution. We will describe how
this relates to problem tractability.

{\bf Motivation for Empirical Mapping.}
A key failure mode of iterative solvers is finding a local minimum of
poor quality.
This problem can be largely avoided
through good initial guesses
(perhaps via a relaxation of the problem).
Even then, it is helpful to have some guidance as to what quality of
guess is necessary. Indeed, for some
optimization problems,
sufficiently good initial guesses are
rare.
In this section we study the problem
experimentally, and then in \Cref{sec:theory} we attempt to explain our
observations theoretically.

We begin by fixing a problem instance. Then we collect
many local minima: we choose $N_{lm}$ initial guesses sampled
uniformly at random in $\SO(3)^n$ and then run a
Levenberg-Marquardt
solver~\cite{ceres-solver} to efficiently iteratively minimize an
$l_p$ cost. This gives us $N_{lm}$
(possibly not distinct) local minima.

\begin{figure*}
    \begin{center}
        \hfill
        \begin{tabular}{m{0.01\linewidth}m{0.98\linewidth}}
            \toprule
            (1) & \includegraphics[trim=0 42 0 38,clip,width=\linewidth]{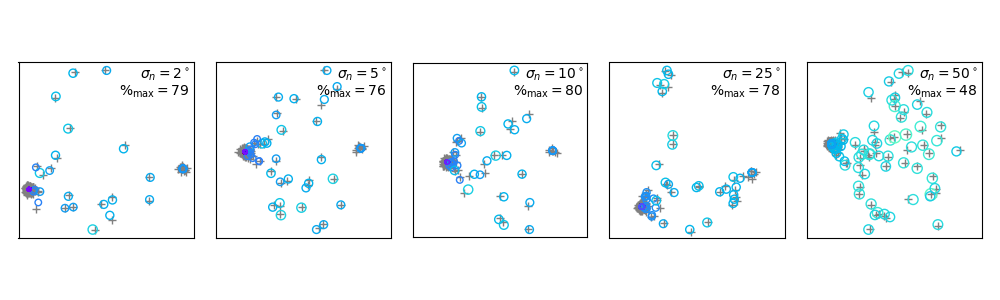} \\
            \midrule[0.2pt]
            (2) & \includegraphics[trim=0 42 0 38,clip,width=\linewidth]{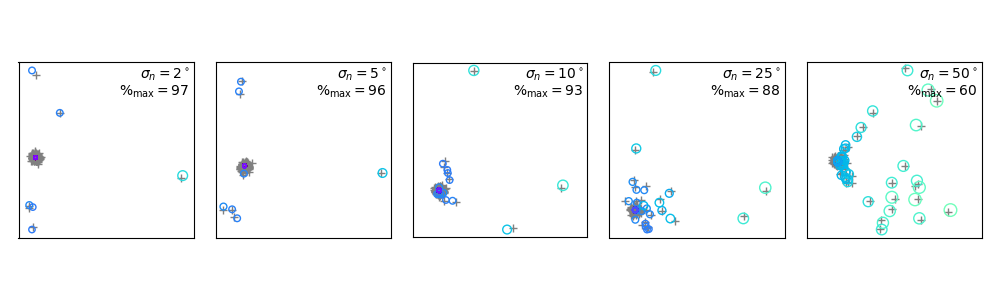} \\
            \midrule[0.2pt]
            (3) & \includegraphics[trim=0 42 0 38,clip,width=\linewidth]{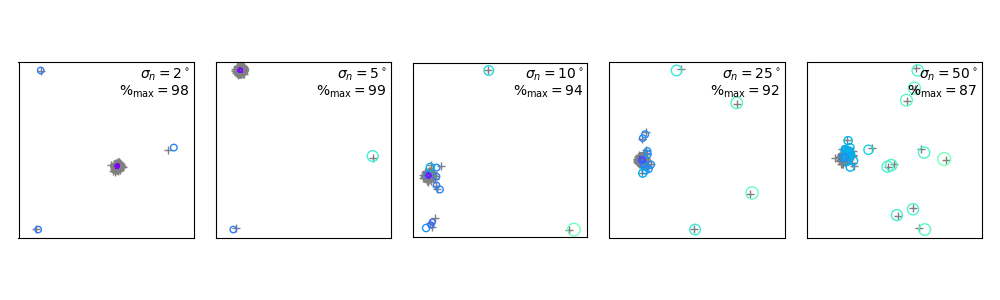} \\
            \midrule[0.2pt]
            (4) & \includegraphics[trim=0 42 0 38,clip,width=\linewidth]{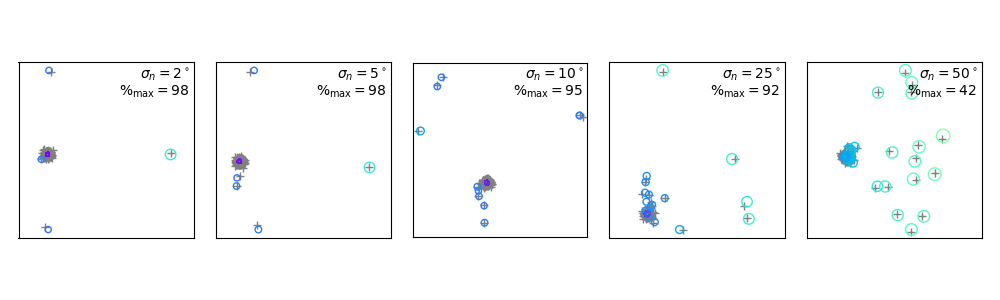} \\
            \midrule[1.0pt]
            (5) & \includegraphics[trim=0 42 0 38,clip,width=\linewidth]{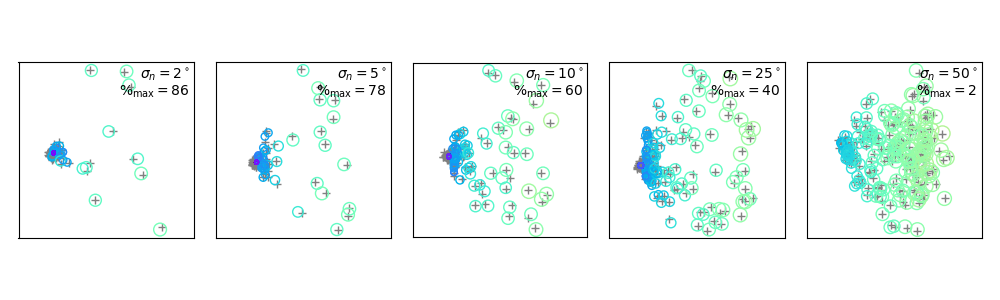} \\
            \midrule[0.2pt]
            (6) & \includegraphics[trim=0 42 0 38,clip,width=\linewidth]{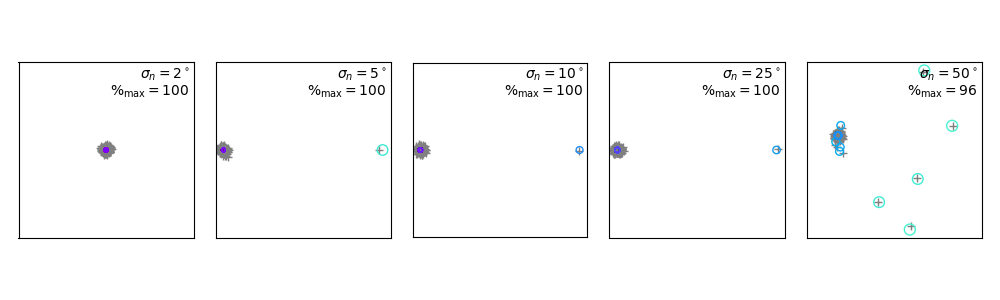} \\
            \bottomrule
        \end{tabular}
        \hfill
    \end{center}
    \caption{2D embeddings of the local minima of rotation averaging problems. Each row is a fixed underlying graph, and the columns are increasing noise levels ($\sigma_n$). Both color and circle radius encode the objective function (cool colors and small radii are low cost). To reveal multiplicity, a small jittered \texttt{+} is plotted in grey behind each point. Additionally, $\%_{\max}$ gives the multiplicity of the most common minimum. Notice that some problem/noise level combinations have essentially one unique dominant local minimum, while others have many. \Cref{tab:empirical_key} describes the graphs for each row.}
    \label{fig:empirical}
\end{figure*}

Now that we have a large, randomly sampled collection of local minima, we visualize their distribution.
Because the local minima live in the search space $\SO(3)^n$, a $3n$-dimensional manifold, we are unable to plot them directly. Instead, we first compute (gauge-aligned) pairwise distances between each of the $N_{lm}$ minima. This effectively gives us a graph with labeled edge lengths.  We use a spectral projection to embed this graph in a 2D plane. These 2D embeddings are show in \cref{fig:empirical} and described in greater detail below.

{\bf Embedding Experimental Details.}
Each row in \cref{fig:empirical} is a series of experiments on the
same graph; see \cref{tab:empirical_key} for details.
To create each problem instance, we select a ground truth solution for
the given graph by choosing elements of $\SO(3)$ uniformly at
random. We then generate noisy edge measurements by synthetically
adding noise (zero-mean and variance $\sigma_n^2$
Gaussian in the tangent space) to
each edge's ground truth relative rotation. The columns
of \cref{fig:empirical} are increasing noise levels. Within a row,
only $\sigma_n$ varies. The graph and ground truth solution are
fixed.

We use $N_{lm}=200$ random starts for each experiment. For the spectral graph embedding, we use the diffusion kernel $d \mapsto e^{-d^2/ \sigma_{d}}$ to transform distances to similarities, where $d$ is the distance between two local minima. We used $\sigma_{d}=\frac{\pi}{4}$ for all of our experiments. The synthetic noise $\sigma_n$ is labeled on each experiment in \cref{fig:empirical}.

\begin{table}[h]
    \footnotesize
    \begin{center}
        \begin{tabular}{lllr}
            \toprule
            \textbf{row} & \textbf{graph parameters} & \textbf{connectivity} & $\alpha_{\max}$~\cite{eriksson-cvpr-18} \\
            \midrule
            & \multicolumn{3}{l}{Watts-Strogatz small world graphs} \\[2pt]
            (1) & $n=40$, $k=16$, $p=0.0$ & $\lambda_2(G)=4.6$ &  14.7\textdegree \\
            (2) & $n=40$, $k=16$, $p=0.2$ & $\lambda_2(G)=7.7$ &  19.0\textdegree \\
            (3) & $n=40$, $k=16$, $p=0.5$ & $\lambda_2(G)=8.2$ &  19.2\textdegree \\
            (4*) & $n=40$, $k=16$, $p=1.0$ & $\lambda_2(G)=9.5$ & 20.2\textdegree \\
            \midrule
            & \multicolumn{3}{l}{$G_{nm}$ random graphs} \\[2pt]
            (5) & $n=40$, $m=200$ & $\lambda_2(G)=4.2$  & 13.6 \textdegree \\
            (4*) & $n=40$, $m=240$ & $\lambda_2(G)=9.5$ & 20.2 \textdegree \\
            (6) & $n=40$, $m=400$ & $\lambda_2(G)=12.0$ & 12.0 \textdegree \\
            \bottomrule
        \end{tabular}
    \end{center}
    \caption{Descriptions of the graphs in \cref{fig:empirical}. The algebraic connectivity $\lambda_2(G)$ features in the results in \cref{sec:theory}. The quantity $\alpha_{\max}$ is from~\cite{eriksson-cvpr-18}: their solution is certifiably optimal if all residuals are less than $\alpha_{\max}$. \emph{\; (*) Note that graph (4) is an instance of both Watts-Strogatz and $G_{nm}$ random graphs when $p=1$.}}
    \label{tab:empirical_key}
\end{table}

{\bf How to Interpret the Embeddings.}
Each small colored circle in the plots is a local minimum. Within each plot, absolute distances are not meaningful, but relative distances within each plot correspond (imperfectly) to distances between minima. Gauge is quotiented out when computing distances, as described in \cref{sec:theory}. The colors and radii of the points are both proportionate to the objective function of the minima (cool colors are lower cost and warm colors are higher).

Jittered points are also plotted behind in light grey to reveal multiplicity. For example, the jitter reveals that of the four unique minima in the plot at row 3, column 1, the minimum with lowest cost is strongly dominant. The dominance of the largest minimum ($\%_{\max}$) is labeled on each experiment. This is the percent of the $N_{lm}$ runs that reached the highest multiplicity minimum.

The qualitative results in \cref{fig:empirical} are stable under repeated trials. Different random graphs and random synthetic noise always result in different 2D embeddings, but all of the trends discussed below are still present.

{\bf Observations and Trends.}
Rows (1) through (4) are Watts-Strogatz small world random graphs~\cite{strogatz-nature-01}. This family of graphs has a rewiring parameter $p$ that smoothly interpolates between a $k$-connected cycle graph and $G_{nm}$, the random graph model that randomly samples from all graphs with exactly $n$ vertices and $m$ edges.

These graphs are interesting to us because $G_{nm}$ graphs (and the closely related $G_{np}$ Erd\H{o}s R\'enyi graphs) are not that representative of real rotation averaging problems. For their size, they have small radius and high algebraic connectivity. By varying the Watts-Strogatz rewiring parameter $p$ we keep the size of the graph fixed but vary the connectivity.
%
%
Furthermore, the Watts-Strogatz model comes closer to capturing a qualitative property
of image graphs: cameras which have similar pose are more likely to overlap, yet dissimilar cameras do occasionally see the same thing (perhaps on a tall or prominent structure).

Row (1) is the least-connected extreme of the Watts-Strogatz family, with the lowest algebraic connectivity (\cref{tab:empirical_key}). Row (4) is the opposite: with $100\%$ rewiring, this is just a $G_{nm}$ random graph with $240$ edges.

Some problem instances look very easy to solve. For instance, in row (3) column 1, 97.5\% of the random initializations were sufficient to find the dominant minimum. It certainly seems plausible that this minimum is in fact the global optimum. If such a high percentage of random initial guesses suffice, then a non-random guess (such as from a relaxed solver) seems very likely to succeed.

In contrast, consider row (1) column 5. This is a much noisier problem. Now there are many distinct local minima, many of which are very close to each other. An iterative solver may have difficulty finding a true global optimum.

Finally, consider the trends in these graphs:
\begin{itemize}[noitemsep]
    \item At low noise levels there are very few distinct local minima. One of these is usually dominant and appears to be the global optimum.
    \item As noise levels increase, more distinct local minima appear, and they get closer to the dominant minimum.
    \item At fixed noise levels, the more connected graphs look easier. They have fewer distinct minima and these are better separated.
    \item Going from rows (1) to (2) to (3), the connectivity goes up and the problems appear to get easier. However, rows (3) and (4) look very similar. The extra connectivity does not appear to have much affect. Is there a critical level of connectivity?
\end{itemize}

Rows (4) - (6) vary a different graph parameter. Recall that the fully rewired Watts-Strogatz graph in (4) is also a $G_{nm}$ random graph with 240 edges. Row (5) is a $G_{nm}$ graph with a few fewer edges, and (6) with a few more. Notice how sharply these changes affect the plots of minima! Row (5), which has fewer edges, appears with only $2^\circ$ of noise to look as messy as row (4) does with $25^\circ$. Row (6) is still well-behaved with a remarkable $25^\circ$ of noise.

{\bf Practical Takeaways.}
These visualizations indicate that some instances of the $l_p$ rotation averaging problem are easy---in the sense that a single minimum dominates the cost surface. This is presumably the global optimum, and almost any initialization is adequate to find it.

On the other hand, some problems are qualitatively different. They have many local minima, and these minima are close together. Furthermore, there may even be minima that are very similar in cost but not close to each other. This raises important questions about whether the problem is well-enough posed for solutions (even the global solution, were it available) to be useful.

In these experiments, for the same number of nodes and edges, more connected graphs were better. Unsurprisingly, for all else equal, lower noise was also better. See the supplemental for a discussion of outlier edges.

Looking at the last column of Table 1, we see that the Eriksson \etal's certification~\cite{eriksson-cvpr-18} is sufficient, but not necessary. The bound behaves as expected on the Watts-Strogatz graphs, but not on the $G_{nm}$ problems. This may be due to fragile dependence on the maximum degree of the graph.

\section{Improved Local Convexity Bounds}
\label{sec:theory}
Now we turn to seeking a theoretical description of the effects shown in
\Cref{sec:empirical}. We proceed by studying the extent of \emph{local
convexity} in a rotation averaging problem instance. We generally follow the
calculations in~\cite{wilson-eccv-16}, but we add two pieces: (1) the
calculations are generalized from $l_2$ to $l_p$ costs, and (2) we describe
the gauge ambiguity in a geometrically natural way, leading to a large
improvement in the bounds.

Convex problems enjoy a prominent place in optimization because they provide a guarantee: all local minima are global minima. For rotation averaging, this helpful property does not hold. However, we can proceed with an analysis using \emph{local convexity}, a far weaker, but closely related property.

\begin{defn}
    A rotation averaging problem $(G, \widetilde{\mathcal{R}})$ is locally convex at a solution $\mathcal{R}$ if there exists an open ball around $\mathcal{R}$ on which the problem is convex.
\end{defn}

Practically speaking, for functions that are at least twice differentiable, local convexity is a statement that the Hessian matrix of the problem is
positive semi-definite.%
\footnote{This will serve for most of our analysis, but when $1 \leq p < 2$ and at a point where a residual is zero we lack differentiability. Instead, we look at the cost function along geodesics parameterized at constant speed within a small ball. We only need consider geodesics which pass exactly through the point. Finally, we note that along such a geodesic the cost function is a scaled version of $f(\theta) = |\theta|^p$, which is known to be convex for $p \geq 1$.}

In the context of rotations averaging, local convexity limits local minima, because if we have local convexity on a convex region, there cannot be distinct local minima there.

In \cite{wilson-eccv-16}, it was shown that the $l_2$ geodesic variant of the rotations averaging cost function frequently exhibits local convexity, and that the extent of that local convexity is bounded by properties of the problem instance. The crux of that argument was an appreciation of the \emph{gauge ambiguity}---the local convexity in the problem was only visible after the gauge had been treated. In this section, we revisit that treatment with a more natural description of the gauge.

\subsection{A more natural perspective on gauge}
By gauge ambiguity, we mean that our cost function is invariant to right-multiplication by any element of $\SO(3)$.
This is a free group action of $\SO(3)$ on $\SO(3)^n$, leading us to consider our optimization problem over the quotient manifold
$\SO(3)^n /\left( \mathcal{R} \sim \mathcal{R}\rot{S}\right)$. (The natural metric on this manifold is $d_\mathrm{quot}(\mathcal{R}_1, \mathcal{R}_2) = \min_{\rot{S} \in \SO(3)} d(\mathcal{R}_1, \mathcal{R}_2 \rot{S})$.)
This quotient no longer has a gauge ambiguity; we have exactly removed the symmetry in question. We would like to analyze the local convexity of this new problem, which raises the question, ``How do derivatives of a function on a manifold relate to derivatives of that function on a quotient manifold?''

\begin{figure}
    \begin{center}
        \includegraphics[width=0.9\linewidth]{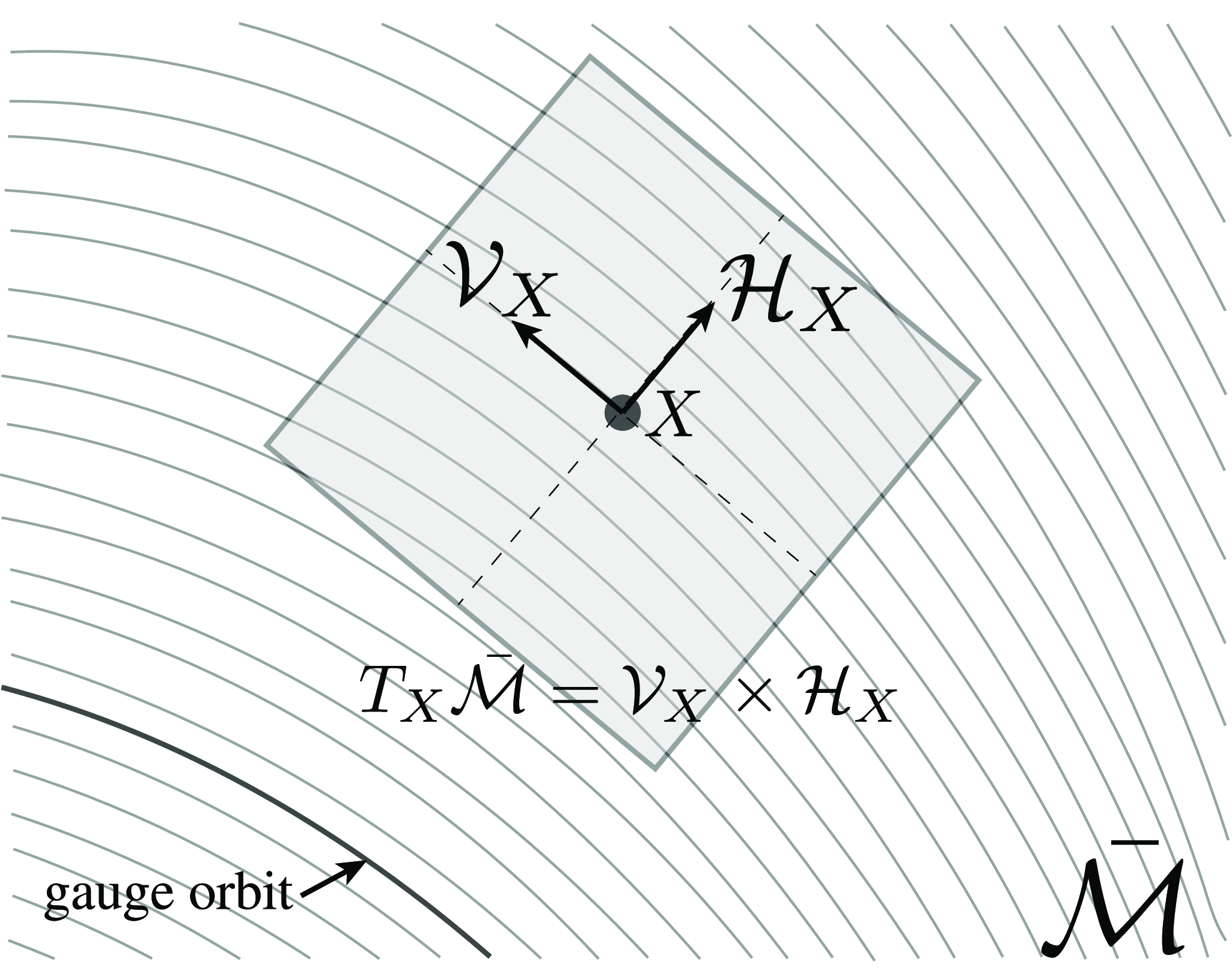}
    \end{center}
    \caption{A picture of a 2D manifold $\bar{\mathcal{M}}$ partitioned into 1D orbits. The tangent space $T_X \bar{\mathcal{M}}$ at any point $X \in \bar{\mathcal{M}}$ can be decomposed as the product of two orthogonal vector spaces: a vertical space (which follows the orbits) and a horizontal space (which is orthogonal to the orbits).}
    \label{fig:tangent_spaces}
\end{figure}

Quotients of Riemannian manifolds have a particularly nice
structure~\cite[Ch.~3]{absil-09}.
Points in the quotient space correspond
to gauge-equivalent orbits. A tangent space $T_X\mathcal{M}$ on the quotient manifold is related
to the tangent space $T_X \bar{\mathcal{M}}$ on
$\bar{\mathcal{M}} = \SO(3)^{n}$ as shown in \cref{fig:tangent_spaces}.
In particular, note that the original tangent space $T_X \bar{\mathcal{M}}$ is a direct product of two orthogonal vector spaces. The vertical space is the subspace consisting of motions that remain in the same gauge orbit, and the horizontal space is the space of motions orthogonal to that. Notably, the horizontal space is identified with the quotient manifold's tangent space: $\mathcal{H}_X = T_X \mathcal{M}$. Furthermore, through this identification, derivatives of functions on quotient manifolds are simply the projections onto $\mathcal{H}$ of derivatives in the original manifold.

%

We can explicitly state what these spaces are for our problem. A perturbation in a tangent space to $\SO(3)^n$ can be expressed as an $n$-tuple $(\mathbf{x_1}, \ldots, \mathbf{x_n})$, where the $\mathbf{x}_i$'s are all in $\mathbb{R}^3$. (Properly speaking, then the tangent space elements are the skew symmetric matrices $[\mathbf{x}_i]_\times$.) Each such tangent space element is mapped through the $\exp$ map to a new element of $\SO(3)^n$:
\[ \exp_\mathcal{R}(\mathbf{x_1}, \ldots, \mathbf{x_n}) =
\left[ \begin{array}{c}
\rot{R_1} \exp [\mathbf{x}_1]_\times \\
\vdots \\
\rot{R_n} \exp [\mathbf{x}_n]_\times \\
\end{array} \right].
\]
If this motion is gauge-preserving we must have
\[ \exp [\mathbf{x}_1]_\times = \ldots = \exp [\mathbf{x}_n]_\times = \rot{S} \]
for some $\rot{S} \in \SO(3)$. The inverse of the $\exp$ map is $\log$, which maps from the manifold into the tangent space. For small $\mathbf{x}$, $\log [\mathbf{x}]_\times$
is one-to-one, so $\mathbf{x}_1
= \mathbf{x}_2 = \ldots = \mathbf{x}_n$.
In other words, in Kronecker product notation, the vertical space is this subspace of $\mathbb{R}^{3n}$:
\begin{equation}
    \mathcal{V} = \Span \left\{
        \left[ \begin{array}{c}1\\0\\0\\ \end{array} \right] \otimes \mathbf{1}_{n} ,
        \left[ \begin{array}{c}0\\1\\0\\ \end{array} \right] \otimes \mathbf{1}_{n} ,
        \left[ \begin{array}{c}0\\0\\1\\ \end{array} \right] \otimes \mathbf{1}_{n}.
    \right\}
\end{equation}
It follows that the horizontal space is $\mathcal{H} = \mathcal{V}^\perp$. Interestingly, $\mathcal{V}$ and $\mathcal{H}$ depend only of $n$. The tangent spaces at every point $\mathcal{R}$ on the manifold share the same description of the vertical and horizontal spaces.

Returning to the original question, we wish to analyze the local convexity of the quotient-manifold version of the $l_p$ rotation averaging problem.
Recall that (given sufficient differentiability) local convexity holds when all eigenvalues of the Hessian matrix are non-negative.
This will hold exactly when the projection of the Hessian matrix onto $\mathcal{V}^\perp$ is positive semi-definite:

%
\begin{thm}
    A rotation averaging problem is locally convex with respect to the $l_p$ geodesic cost function when
    \begin{equation}
        \min_{\mathbf{x} \in \mathbf{V}^\perp, \|\mathbf{x}\| = 1} \mathbf{x}^\top \mathbf{H} \mathbf{x} > 0,
    \end{equation}
    where $\mathbf{H}$ is the Hessian of the $l_p$ cost function (on $\SO(3)^n$; see below).
\end{thm}

\subsection{Some Lower-Bound Approximations}
Now that we have a better way to factor the gauge ambiguity out of our local convexity analysis, we begin with the Hessian matrix computed in~\cite{wilson-eccv-16} and follow it through two approximations.
The smallest eigenvalue of the (gauge-projected) Hessian matrix describes how the graph structure of the problem interacts with the magnitude and directions of residuals, determining local convexity.
Unfortunately, this description is complicated. In this section we aim to get insight through sacrificing some accuracy for interpretability. In each case, our new gauge approach yields superior bounds over analogous prior results~\cite{wilson-eccv-16}.

{\bf The Structure of the Hessian.}
We begin by giving the exact Hessian for the unquotiented $l_p$ cost function, as derived by following the chain rule calculations in~\cite{wilson-eccv-16} for $p>1$. Recall that the cost function (\cref{eqn:problem}) is a sum of terms on each edge in the graph. Likewise, the Hessian is a sum of terms for each edge in the cost function:
\begin{equation}
    \mathbf{H} = \sum_{(i,j) \in E} \mathbf{H}_{ij}.
\end{equation}
where the terms are $3n$-by-$3n$ matrices composed of 3-by-3 blocks. For each term, exactly four 3-by-3 blocks (corresponding to the $i$th and $j$th block rows and columns) are non-zero. These blocks are expressed in terms of the residuals $\rot{R}_i^\top \widetilde{\rot{R}}_{ij} \rot{R}_j$ on each edge. We write these residuals in their tangent spaces as $[\theta_{ij} \hat{\mathbf{w}}_{ij}]_\times = \log \left( \rot{R}_i^\top \widetilde{\rot{R}}_{ij} \rot{R}_j \right) $.
So, where $\theta_{ij}$ is the magnitude of the residual rotation on edge $(i,j)$, and $\mathbf{w}_{ij}$ is its axis of rotation.
\begin{equation}
    \small
    \renewcommand{\kbldelim}{[}
    \renewcommand{\kbrdelim}{]}
    \mathbf{H}_{ij} =
        \kbordermatrix{ & & i & & j & \cr
              & & \vdots & & \vdots & \cr
            i &\ldots & \mathbf{S}_{ij} & \ldots & -\mathbf{S}_{ij} + \mathbf{A}_{ij}^\top & \ldots \cr
              & & \vdots & & \vdots & \cr
            j & \ldots & -\mathbf{S}_{ij} + \mathbf{A}_{ij} & \ldots & \mathbf{S}_{ij} & \ldots \cr
              & & \vdots & & \vdots & \cr
    },
\end{equation}
where the 3-by-3 blocks making up each $\mathbf{H}_{ij}$ are
\begin{align*}
    \mathbf{S}_{ij} &= p(p-1)\theta_{ij}^{p-2}\mathbf{w}_{ij} \mathbf{w}_{ij}^\top +
    \frac{p}{2}\theta_{ij}^{p-1}(\mathbf{I}_3 - \mathbf{w}_{ij} \mathbf{w}_{ij}^\top), \\
    \mathbf{A}_{ij} &= \frac{p}{2}\theta_{ij}^{p-1} [\mathbf{w}_{ij}]_\times.
\end{align*}
Notice that the symmetric $\mathbf{S}_{ij}$ terms of the Hessian are arranged in a graph-Laplacian like structure, but the antisymmetric $\mathbf{A}_{ij}$ terms are not.

{\bf An Isotropic Bound.}
A first approach to taming
this Hessian is to
find a bound that depends
only on residual magnitudes.
The best isotropic (direction independent) lower bound for the $\mathbf{S}_{ij}$ terms is
\begin{equation}
    \mathbf{S}_{ij} \succ  \min\left(\frac{p}{2}\theta_{ij}^{p-1}, p(p-1)\theta_{ij}^{p-2} - \frac{p}{2}\theta_{ij}^{p-1} \right)\mathbf{I}_3.
\end{equation}
Let's call those values $\alpha_{ij}$, so that we have $\mathbf{S}_{ij} \succ \alpha_{ij} \mathbf{I}_3$. Notice that $\alpha_{ij}$ is purely a convenience function of $\theta_{ij}$.
Now, the indefinite skew contribution $\mathbf{S}_{ij}$ can be isotropically bounded above as a block:
\begin{equation}
    \left[ \begin{array}{cc}
            \mathbf{0} & \mathbf{A}_{ij}^\top \\
            \mathbf{A}_{ij} & \mathbf{0} \\
    \end{array} \right]
    \prec
    \frac{p}{2}\theta_{ij}^{p-1}
    \left[ \begin{array}{cc}
            \mathbf{I}_3 & \mathbf{0} \\
            \mathbf{0} & \mathbf{I}_3 \\
    \end{array}\right].
\end{equation}
Continuing
as in~\cite{wilson-eccv-16},
we see that for a single Hessian term,
\begin{equation}
    \mathbf{H}_{ij} \succ \left(
    \underbrace{\left[ \begin{array}{cc}
            \alpha_{ij} & -\alpha_{ij} \\
            -\alpha_{ij} & \alpha_{ij} \\
    \end{array} \right]}_{\text{PSD Laplacian term}}
    -\underbrace{\frac{p}{2}\theta_{ij}^{p-1}
    \left[ \begin{array}{cc}
            \mathbf{I}_3 & \mathbf{0} \\
            \mathbf{0} & \mathbf{I}_3 \\
    \end{array}\right]}_{\text{non-PSD curvature term}}
    \right) \otimes \mathbf{I}_3.
\end{equation}
Considering all the residual terms together, we will write this as
$\mathbf{H} \succ \left( \mathbf{L}(\alpha_{ij}) - \mathbf{D}(\theta_{ij}) \right) \otimes \mathbf{I}_3$.
$\mathbf{L}$ is a standard graph Laplacian matrix~\cite{luxberg-07}
with edge weights $\alpha_{ij}$, and $\mathbf{D}$ is a diagonal degree
matrix with elements:
\begin{equation}
    \mathbf{D}_{ii} = \sum_{j | (i,j) \in E} \frac{p}{2}\theta_{ij}^{p-1}.
\end{equation}
Multiplying by $\mathbf{D}^{-\frac{1}{2}}$ on both sides, we see that
this is sufficient for $\mathbf{H} \succ \mathbf{0}$:
\begin{equation}
    \mathbf{L}_{\text{norm}} = \left( \mathbf{D}(\theta_{ij})^{-\frac{1}{2}} \mathbf{L}(\alpha_{ij}) \mathbf{D}(\theta_{ij})^{-\frac{1}{2}} \right) \succ \mathbf{I}_n.
\end{equation}
The smallest eigenvalue of a graph Laplacian is 0, so this condition appears to be unsatisfiable. We must use what we know about gauge to proceed. Recall that considering the problem from a quotient manifold perspective effectively meant projecting the tangent space off of the vertical space and onto the horizontal space. This means that (allowing for the Kronecker product), we require:
\begin{equation}
    \min_{\mathbf{x} \perp \mathbf{1}_n, \|\mathbf{x}\| = 1} \mathbf{x}^\top \mathbf{L}_{\text{norm}} \mathbf{x} > 1.
\end{equation}
That is, $\mathbf{L}_{\text{norm}}$ is an particular weighted, normalized graph Laplacian. It represents the interaction of graph topology and residual noisiness (but not residual directions, which we approximated away), it bounds local convexity. Projecting this matrix off of the gauge vector $\mathbf{1}_n$ represents considering the problem as being solved on a quotient manifold.

{\bf An Even Simpler Bound.}
If we are a bit more destructive, we can reach a bound that cleanly separates structure and noise terms. Let's begin back with this sufficient condition for local convexity:
\begin{equation}
    \mathbf{L}(\alpha_{ij}) \succ \mathbf{D}(\theta_{ij}).
\end{equation}
This next condition is still sufficient, but much weaker:
\begin{equation}
    \lambda_{\min}(\mathbf{L}(\alpha_{ij})) > \lambda_{\max}(\mathbf{D}(\theta_{ij})).
\end{equation}
Now, the eigenvector corresponding to the zero eigenvalue of a graph Laplacian is $\mathbf{1}_n$, the vector of all ones. This means that (allowing for the Kronecker product) the gauge subspace corresponds perfectly with the 0-eigenvector of the graph Laplacian. Note that the largest value of a diagonal matrix is its largest entry, so we have a bound:
\begin{equation}
    \lambda_2(\mathbf{L}(\alpha_{ij})) > \max_{i \in V} \sum_{j | (i,j) \in E} \frac{p}{2}\theta_{ij}^{p-1}.
\end{equation}
Note that in the $p=2$ case, the right hand side is simply the maximum weighted degree of the residual graph. If we wish to bound even more aggressively, we can bound by the smallest $\alpha$, giving:
\begin{equation}
    \underbrace{\lambda_2(\mathbf{L})}_{structure} > \underbrace{\frac{\max_{i \in V} \sum_{(i,j) \in E} \frac{p}{2}\theta_{ij}^{p-1}}{\min_{ij} \alpha_{ij}}}_{residuals}.
    \label{eqn:final_bound}
\end{equation}
This gives a perfect separation of the condition for local convexity into
a graph structure term (the so-called \emph{algebraic connectivity})
a term that depends on residual noise levels. Notice that this bound differs from the one given in~\cite{wilson-eccv-16} in that a factor of $n$ has disappeared from the denominator of the structure term in \cref{eqn:final_bound}. This is solely a consequence of the improved gauge treatment substantially increases the practical utility of the bound.

\section{Application and Conclusion}
\label{sec:conclusion}
In this paper we studied the cost surface of the geodesic $l_p$ rotation
averaging problem. We gave empirical evidence that changes in problem structure
and noise can cause sharp transitions between easy and challenging distributions
of local minima. We then derived a bound on this behavior: connectivity must
balance noise to get a single dominant minimum. These bounds are an
$n$-fold improvement over prior work, achieved through treating the problem's
gauge ambiguity in a geometrically natural way.

We see this analysis being useful for the development of new intrinsic-metric
rotation averaging solvers. Our theoretical work shows how to remain in the easy regime. We hope that this analysis can provide
guidance in pipeline design for forming rotation averaging problem instances.

There is a particular application to
hierarchical / merging-based solver schemes. In these approaches (which are
a current area of active study) small subproblems are selected, solved, and
merged into a larger solution. Our results are especially applicable here
because unlike in single-shot schemes where the problem is provided as-is,
merging solvers have free choice of
their subproblems. Schemes that can maintain elevated connectivity may do
better at avoiding local minima traps.

\smallskip
\noindent
\textbf{Acknowledgements.} This work was supported in part by the National Science Foundation under DMS-1620038.

{\small
\bibliographystyle{ieee_fullname}
\bibliography{references}
}

\end{document}


\title{Supplemental Material for:  \\ On the Distribution of Minima in Intrinsic-Metric Rotation Averaging}

\author{Kyle Wilson\\
Washington College\\ Department of Math and Computer Science \\
Chestertown, MD\\
{\tt\small kwilson24@washcoll.edu}
\and
David Bindel\\
Cornell University \\ Department of Computer Science\\
Ithaca, NY\\
{\tt\small bindel@cs.cornell.edu}
}

\maketitle
\thispagestyle{empty}

\section{Connecting the Empirical and Theory Contributions}
In the theory sections of this paper we identified the two salient properties that affect local convexity:
\begin{description}
    \item[Noise:] Lower edge noise yields more local convexity
    \item[Graph Structure:] Greater connectivity yields more local convexity.
\end{description}
The interaction of the two factors is complicated, and we an give exact expression in terms of the spectrum of a particular weighted graph Laplacian (Equation 5). However, we also separate the two factors to seek insight. As shown in Equation 17, algebraic connectivity ($\lambda_2(G)$, the second-smallest eigenvalue of a graph's Laplacian matrix) is a useful measure of the graph structure's contribution to local convexity.

\begin{figure}
    \begin{center}
        \includegraphics[width=0.90\linewidth]{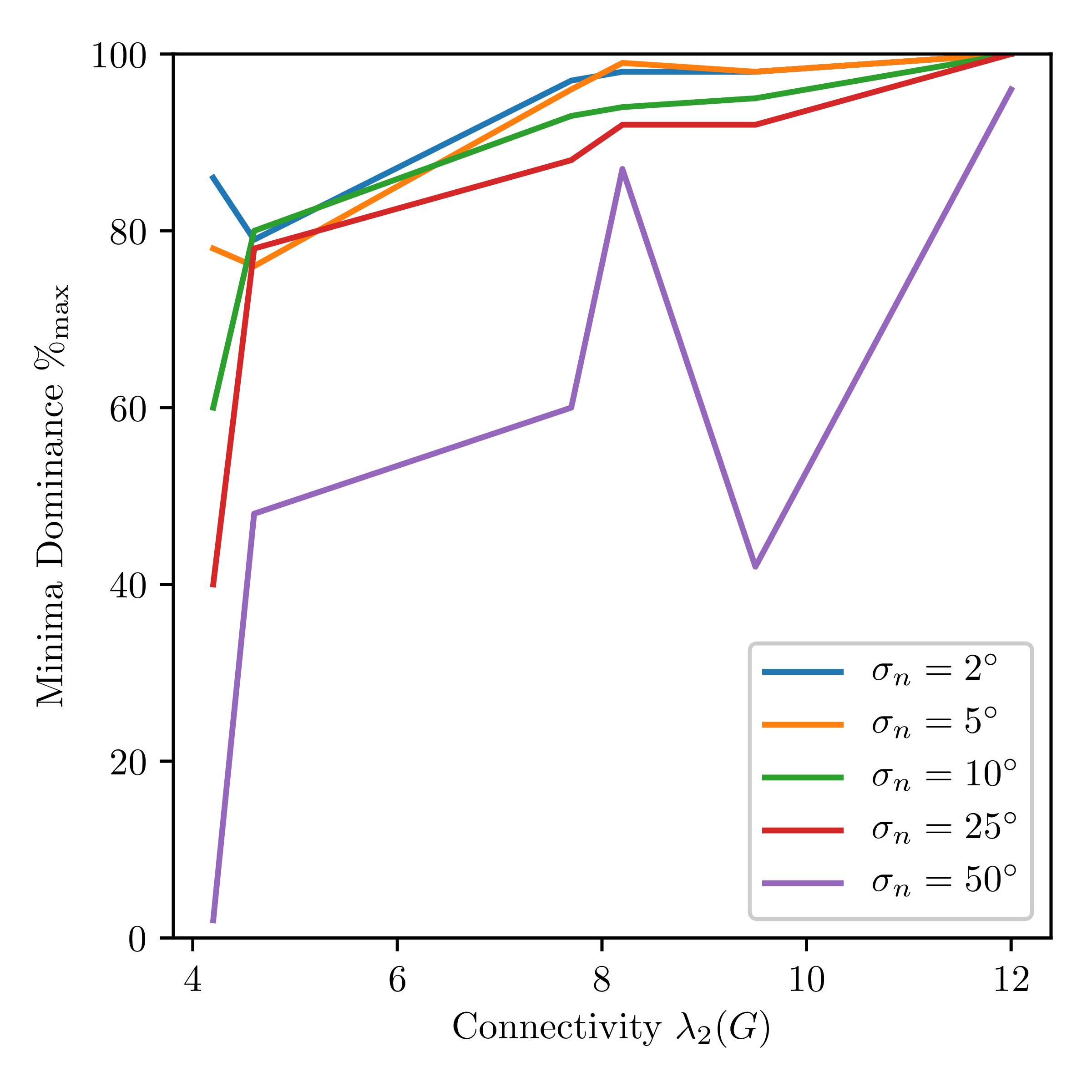}
    \end{center}
    \caption{A plot showing how problem difficulty, as quantified by $\%_{\max}$, the multiplicity of the most common minimum, is affected by noise level and problem connectivity.}
    \label{fig:conn_vs_dom}
\end{figure}

In \cref{fig:conn_vs_dom} we look at noise and connectivity on the empirical experiments from Section 4. We visualize three factors:
\begin{itemize}
    \item Synthetic noise level $\sigma_n$
    \item Graph connectivity $\lambda_2(G)$
    \item Multiplicity of the most common minimum $\%_{\max}$
\end{itemize}
For easy problem instances there will be a single dominant local minimum. Nearly 100\% of all of the minima we found will be identical.

Notice the following in \cref{fig:conn_vs_dom}:
\begin{itemize}
    \item Lower $\lambda_2(G)$ tends to yield lower $\%_{\max}$.
    \item Higher $\sigma_n$ tends to yield lower $\%_{\max}$.
    \item The $\%_{\max}$ is sharply lower below a certain connectivity and below a certain noise level.
\end{itemize}

\section{Empirical Experiments with Outliers}
The empirical experiments in Section 4 all used a Gaussian-like noise model. However, real problems instances commonly have heavy-tailed noise. Here we present the same types of experiments as in Section 4 but with an inlier/outlier noise model.

\begin{figure*}
    \begin{center}
        \begin{tabular}{m{0.01\linewidth}m{0.98\linewidth}}
            \toprule
            (1) & \includegraphics[trim=0 42 0 38,clip,width=\linewidth]{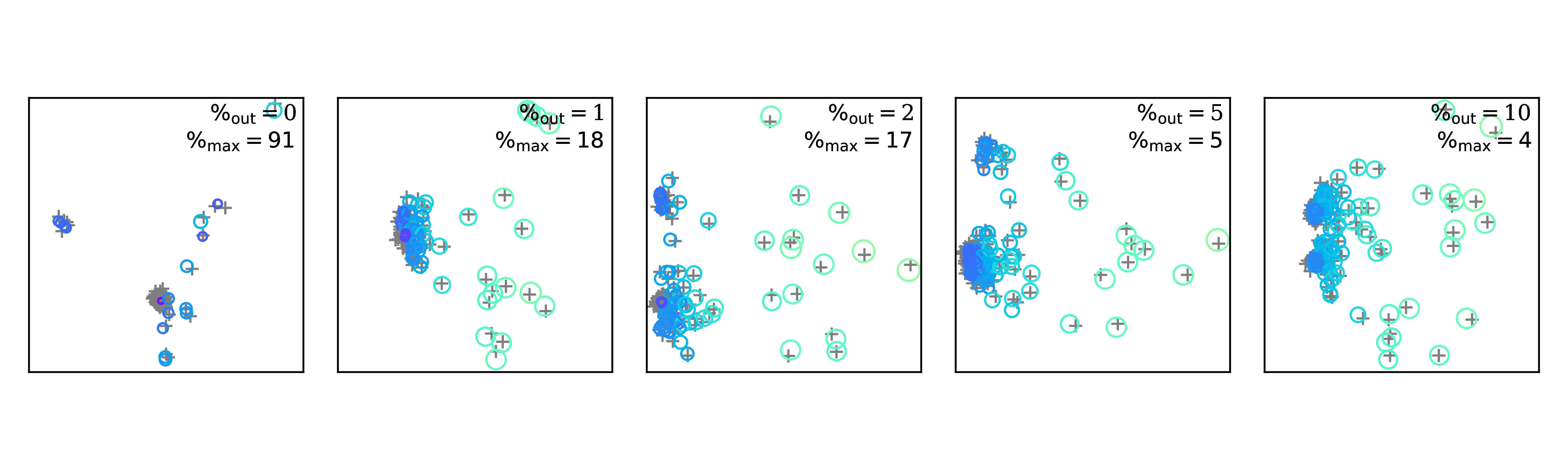} \\
            \midrule[0.2pt]
            (2) & \includegraphics[trim=0 42 0 38,clip,width=\linewidth]{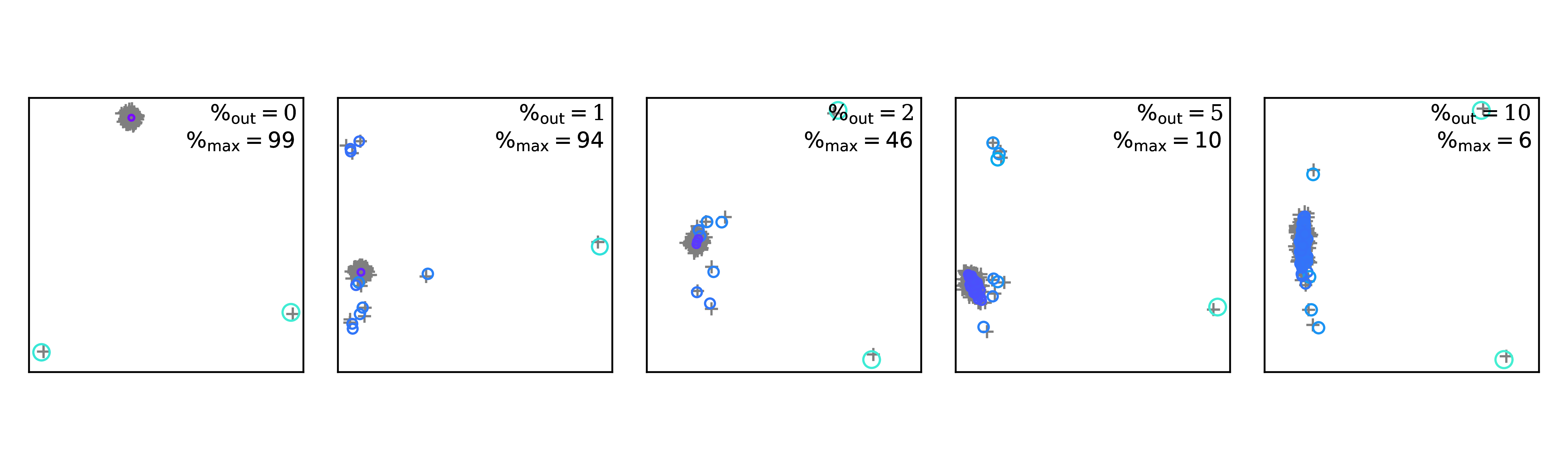} \\
            \bottomrule
        \end{tabular}
    \end{center}
    \caption{Visualizations in the style of Figure 1 from Section 4 of the paper. All problem instances were constructed from the same $G_{nm}$ random graph instance where $n=40$. Row (1) has $m=240$ and row (2) has $m=400$. All instances have the same $5^\circ$ inlier noise applied. The columns are in order of increasing outlier percentage. Even a few outliers introduce many bad minima to the cost surface. Row (2), the better connected graph, tolerates more outliers.}
    \label{fig:outliers}
\end{figure*}

\Cref{fig:outliers} is constructed analogously to rows (4) and (6) of Table 1. All problem instances were made with identical $\sigma_n=5^\circ$ inlier noise and the columns are varying percentages of outlier edges. Outliers are distributed uniformly at random over $\SO(3)$.

Notice that the better connected graph (row 2) tolerates more outliers before bad local minima appear everywhere, but by even 5\% outliers both problems look very hard.

